\documentclass[10pt,twocolumn,letterpaper]{article}

\usepackage{iccv}
\usepackage{times}
\usepackage{epsfig}
\usepackage{graphicx}
\usepackage{amsmath}
\usepackage{amssymb}
\usepackage{multirow}
\usepackage{algorithm}
\usepackage{algpseudocode}

\usepackage{amsmath,amsfonts,bm}

\def\vw{{\bm{w}}}

\def\mH{{\bm{H}}}

\def\gL{{\mathcal{L}}}

\def\gR{{\mathcal{R}}}

\def\gU{{\mathcal{U}}}

\def\Figref#1{Figure~\ref{#1}}
\def\Tabref#1{Table~\ref{#1}}
\def\Secref#1{Section~\ref{#1}}
\def\Algref#1{Algorithm~\ref{#1}}
\def\Eqref#1{{Eq.~\eqref{#1}}}
\def\NAME{{{TS$^{3}$}}}

\usepackage{etoolbox}
\makeatletter
\patchcmd{\maketitle}
 {\def\@makefnmark}
 {\def\@makefnmark{}\def\useless@macro}
 {}{}
\makeatother

\usepackage[breaklinks=true,bookmarks=false]{hyperref}

\iccvfinalcopy 

\def\iccvPaperID{****} 

\begin{document}

\title{Teacher Supervises Students How to Learn\\From Partially Labeled Images for Facial Landmark Detection}

\author{Xuanyi Dong$^{1,2}$\thanks{This work was accepted to IEEE ICCV 2019.} and Yi Yang$^{1,3}$\\
$^{1}$SUSTech-UTS Joint Centre of CIS, Southern China University of Science and Technology\\$^{2}$ Baidu Research, $^{3}$ ReLER, University of Technology Sydney\\
{\tt\small xuanyi.dong@student.uts.edu.au},~{\tt\small yi.yang@uts.edu.au}
}

\maketitle

\begin{abstract}
Facial landmark detection aims to localize the anatomically defined points of human faces. In this paper, we study facial landmark detection from partially labeled facial images. A typical approach is to (1) train a detector on the labeled images; (2) generate new training samples using this detector's prediction as pseudo labels of unlabeled images; (3) retrain the detector on the labeled samples and partial pseudo labeled samples. In this way, the detector can learn from both labeled and unlabeled data to become robust.

In this paper, we propose an interaction mechanism between a teacher and two students to generate more reliable pseudo labels for unlabeled data, which are beneficial to semi-supervised facial landmark detection. Specifically, the two students are instantiated as dual detectors.
The teacher learns to judge the quality of the pseudo labels generated by the students and filter out unqualified samples before the retraining stage.
In this way, the student detectors get feedback from their teacher and are retrained by premium data generated by itself.
Since the two students are trained by different samples, a combination of their predictions will be more robust as the final prediction compared to either prediction.
Extensive experiments on 300-W and AFLW benchmarks show that the interactions between teacher and students contribute to better utilization of the unlabeled data and achieves state-of-the-art performance.
\end{abstract}

\section{Introduction}

Facial landmark detection aims to find some pre-defined anatomical keypoints of human faces~\cite{xiong2013supervised,lv2017deep,xiao2017recurrent,tang2018quantized}.
These keypoints include the corners of a mouth, the boundary of eyes, the tip of a nose, {etc}~\cite{shen2015first,sagonas2013300,koestinger2011annotated}.
It is usually a prerequisite of a large number of computer vision tasks~\cite{liu2018exploring,thies2016face2face,blanz2003face}.
For example, facial landmark coordinates are required to align faces to ease the visualization for users when people would like to sort their faces by time and see the changes over time~\cite{dong2018san}.
Other examples include face morphing~\cite{blanz2003face}, face replacement~\cite{thies2016face2face}, etc.

\begin{figure}[!t]
\center\label{fig:high-level}
\includegraphics[width=\columnwidth]{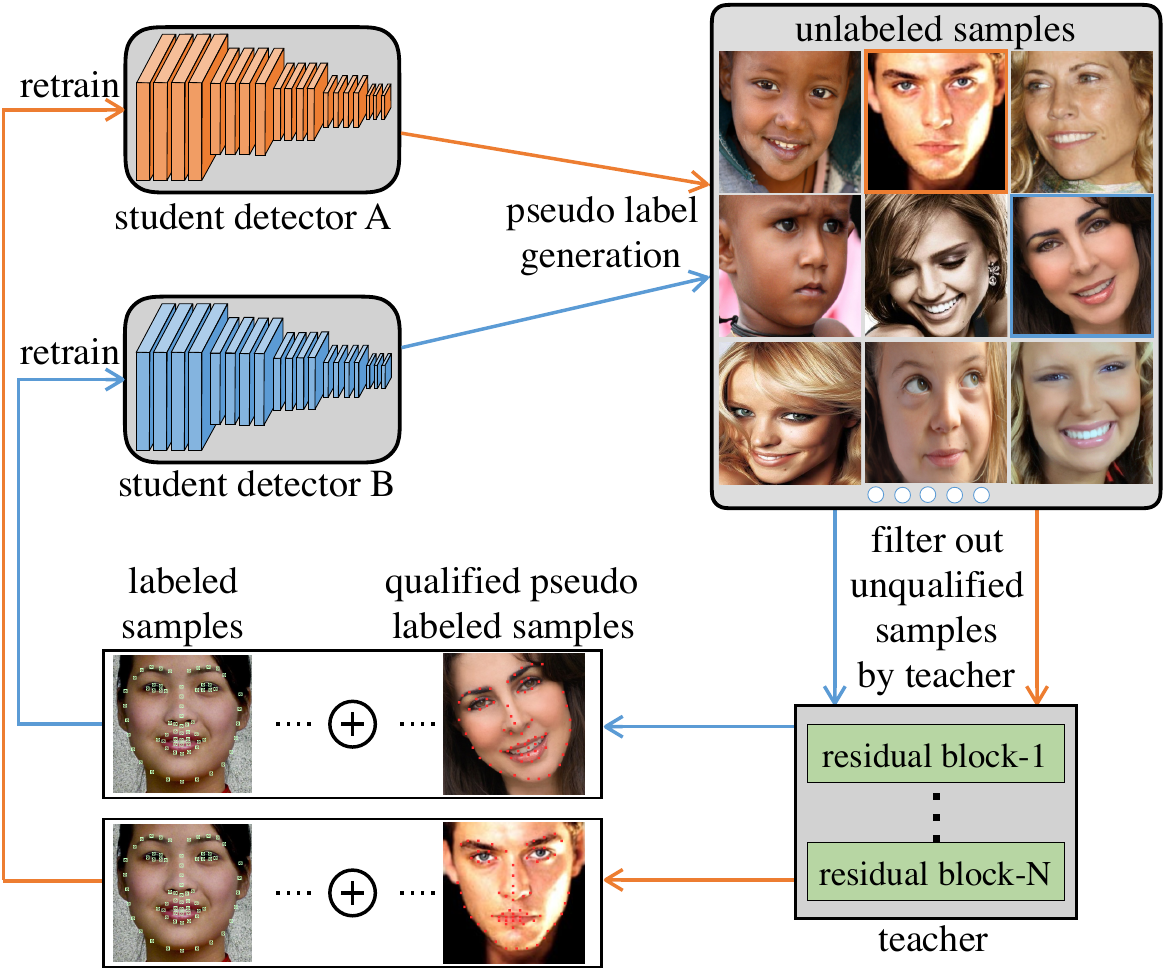}
\caption{
The interaction mechanism between teacher and students. 
Two student detectors learn to generate pseudo labels for unlabeled samples, among which qualified samples are selected by the teacher. These premium pseudo labeled data along with real labeled data is used for the retraining of the students detectors.
}
\end{figure}

The main challenge in recent landmark detection literatures is how to obtain abundant facial landmark labels.
The annotation challenge comes from two perspectives.
First, a large number of keypoints are required for a single face image, e.g., 68 keypoints for each face in the 300-W dataset~\cite{sagonas2013300}.
To precisely depict the facial features for a whole dataset, millions of keypoints are usually required. 
Second, different annotators have a semantic gap.
There is no universal standard for the annotation of the keypoints, so different annotators give different positions for the same keypoints.
A typical way to reduce such semantic deviations among various annotators is to merge the labels from several annotators.
This will further increase the costs of the whole annotation work.

Semi-supervised landmark detection can to some extent alleviate the expensive and sophisticated annotations by utilizing the unlabeled images.
Typical approaches~\cite{jiang2015self,bengio2009curriculum,kumar2010self} for semi-supervised learning use self-training or similar paradigms to utilize the unlabeled samples.
For example, the authors of~\cite{kumar2010self,jiang2015self,ma2017self} adopt a heuristic unsupervised criterion to select the pseudo labeled data for the retraining procedure. 
This criterion is the loss of each pseudo labeled data, where its predicted pseudo label is treated as the ground truth to calculate the loss~\cite{jiang2015self,ma2017self}.
Since no extra supervision is given to train the criterion function, this unsupervised loss criterion has a high possibility of passing inaccurate pseudo labeled data to the retraining stage.
In this way, these inaccurate data will mislead the optimization of the detector and make it easier to trap into a local minimum.
A straightforward solution to this problem is to use multiple models and regularize each other by the co-training strategy~\cite{blum1998combining}.
Unfortunately, even if co-training performs well in simple tasks such as classification~\cite{blum1998combining,ma2017self}, in more complex scenarios such as detection, co-training requires extremely sophisticated design and careful tuning of many additional hyper-parameters~\cite{dong2018few}, e.g., more than 10 hyper-parameters for three models in~\cite{ma2017self}.

To better utilize the pseudo labeled data as well as avoid the complicated model tuning for landmark detection, we propose Teacher Supervises StudentS (TS${^{3}}$).
As illustrated in \Figref{fig:high-level}, TS${^{3}}$ is an interaction mechanism between one teacher network and two (or multiple) student networks.
Two student detection networks learn to generate pseudo labels for unlabeled images.
The teacher network learns to judge the quality of the pseudo labels generated from students.
Consequently, the teacher can select qualified pseudo labeled samples and use them to retrain the students.
TS${^{3}}$ applies these steps in an iterative manner, where students gradually become more robust, and the teacher is adaptively updated with the improved students.
Besides, two students can also encourage each other to advance their performances in two ways.
First, predictions from two students can be ensembled to further improve the quality of pseudo labels.
Second, two students can regularize each other by training on different samples.
The interactions between the teacher and students as well as the students themselves help to provide more accurate pseudo labeled samples for retraining and the model does not need careful hyper-parameter tuning.

To highlight our contribution, we propose an easy-to-train interaction mechanism between teacher and students (TS${^{3}}$) to provide more reliable pseudo labeled samples in semi-supervised facial landmark detection.
To validate the performance of our TS${^{3}}$, we do experiments on 300-W, 300-VW, and AFLW benchmarks.
TS${^{3}}$ achieves state-of-the-art semi-supervised performance on all three benchmarks. 
In addition, using only 30\% labels, our TS${^{3}}$ achieves competitive results compared to supervised methods using all labels on 300-W and AFLW.

\section{Related Work}\label{sec:related-work}

We will first introduce some supervised facial landmark algorithms in \Secref{sec:relate-supervised}.
Then, we will compare our algorithm with semi-supervised learning algorithms and semi-supervised facial landmark algorithm in \Secref{sec:relate-semi}.
Lastly, we explain our algorithm in a meta learning perspective in \Secref{sec:relate-meta}.

\subsection{Supervised Facial Landmark Detection}\label{sec:relate-supervised}

Supervised facial landmark detection algorithms can be categorized into linear regression based methods~\cite{xiong2013supervised,cao2014face} and heatmap regression based methods~\cite{wei2016convolutional,dong2018sbr,dong2018san,newell2016stacked}.
Linear regression based methods learn a function that maps the input face image to the normalized landmark coordinates~\cite{xiong2013supervised,cao2014face}.
Heatmap regression based methods produce one heatmap for each landmark, where the coordinate is the location of the highest response on this heatmap~\cite{wei2016convolutional,dong2018sbr,dong2018san,newell2016stacked,bulat2016convolutional}.
All above algorithms can be readily integrated into our framework, serving as different student detectors.

These supervised algorithms require a large amount of data to train deep neural networks. However, it is tedious to annotate the precise facial landmarks, which need to average different annotations from multiple different annotators. Therefore, to reduce the annotation cost, it is necessary to investigate the semi-supervised facial landmark detection.

\subsection{Semi-supervised Facial Landmark Detection}\label{sec:relate-semi}

Some early semi-supervised learning algorithms are difficult to handle large scale datasets due to the high complexity~\cite{chapelle2006semi}.
Others exploit pseudo-labels of unlabeled data in the semi-supervised scenario~\cite{bachmannips2014learning,bengio2009curriculum,kumar2010self,ma2017self}.
Since most of these algorithms studied their effect on small-scale datasets~\cite{chapelle2006semi,bachmannips2014learning,kumar2010self,ma2017self}, a question remains open: can they be used to improve large-scale semi-supervised landmark detection?
In addition, those self-training or co-training approaches~\cite{kumar2010self,ma2017self,dong2018few} simply leverage the confidence score or an unsupervised loss to select qualified samples.
For example, Dong~et~al.~\cite{dong2018few} proposed a model communication mechanism to select reliable pseudo labeled samples based on loss and score.
However, such selection criterion does not reflect the real quality of a pseudo labeled sample.
In contrast, our teacher directly learns to model the quality, and selected samples are thus more reliable.

There are only few of researchers study the semi-supervised facial landmark detection algorithms.
A recent work \cite{honari2018improving} presented two techniques to improve landmark localization from partially annotated face images.
The first technique is to jointly train facial landmark network with an attribute network, which predicts the emotion, head pose, etc.
In this multi-task framework, the gradient from the attribute network can benefit the landmark prediction.
The second technique is a kind of supervision without the need of manual labels, which enables the transformation invariant of landmark prediction.
Compared to using the supervision from transformation, our approach leverages a progressive paradigm to learn facial shape information from unlabeled data.
In this way, our approach is orthogonal to \cite{honari2018improving}, and these two techniques can complement our approach to further boost the performance.

Radosavovic~et~al.~\cite{radosavovic2018data} applied the data augmentation to improve the quality of generated pseudo landmark labels.
For an unlabeled image, they ensemble predictions from multiple transformations, such as flipping and rotation. This strategy can also be used to improve the accuracy of our pseudo labels and complement our approach. Since the data augmentation is not the focus of this paper, we did not apply their algorithms in our approach.
Dong~et~al.~\cite{dong2018sbr} proposed a self-supervised loss by exploiting the temporal consistence on unlabeled videos to enhance the detector. This is a video-based approach and not the focus of our work. Therefore, we do not discuss more with those video-based approach~\cite{khan2017synergy,dong2018sbr}.

\subsection{Meta Learning}\label{sec:relate-meta}

In a meta learning perspective, our TS$^{3}$ learns a teacher network to learn which pseudo labeled samples are helpful to train student detectors.
In this sense, we are related to some recent literature in ``learning to learn''~\cite{liu2019ppn,ren2018learning,fan2018learning,xu2019autoloss}.
For example, Ren~et~al.~\cite{ren2018learning} learn to re-weight samples based on gradients of a model on the clean validation set.
Xu~et~al.~\cite{xu2019autoloss} suggest using meta-learning to tune the optimization schedule of alternative optimization problems.
Jiang~et~al.~\cite{jiang2018mentornet} propose an architecture to learn data-driven curriculum on corrupted labels.
Fan~et~al.~\cite{fan2018learning} leverage reinforcement learning to learn a policy to select good training samples for a single student model.
These algorithms are designed in the supervised scenarios and can not easily be modified in semi-supervised scenario.

\textbf{Difference with other teacher-student frameworks and generative adversarial networks (GAN).}
Our {\NAME} learns to utilize the output (pseudo labels) of the student model qualified by the teacher model to do semi-supervised learning.
Other teacher-student methods~\cite{tarvainen2017mean,hinton2014distilling,dong2019tas,lee2018teacher} aim to fit the output of the student model to that of the teacher model.
The student and teacher in our work do similar jobs as the generator and discriminator in GAN~\cite{goodfellow2014generative}, while we aim to predict/generate qualified pseudo labels in semi-supervised learning using a different training strategy.

\section{Methodology}\label{sec:method}

In this section, we will first introduce the scenario of the semi-supervised facial landmark detection in \Secref{sec:method-setting}. We explain how to design our student detectors and the teacher network in \Secref{sec:teacher-student}. Lastly, we demonstrate our overall algorithm in \Secref{sec:method-algo}.

\subsection{The Semi-Supervised Scenario}\label{sec:method-setting}

We introduce some necessary notations for the presentation of the proposed method.
Let $\gL = \{ (x_1, y_1), (x_2, y_2), ..., (x_{n_{l}}, y_{n_{l}}) \}$ be the labeled data in the training set and $\gU = \{ (x_{n_l + 1}), (x_{n_l + 2}), ..., (x_{n_{l}+n_{u}}) \}$ be the unlabeled data in the training set, where $x_i$ denotes the $i$-th image, and $y_i \in \gR^{2\times K}$ denotes the ground-truth landmark label of $x_i$.
$K$ is the number of the facial landmarks, and the $k$-th column of $y_i$ indicates the coordinate of the $k$-th landmark. 
$n_{l}$ and $n_{u}$ denote the number of labeled data and unlabeled data, respectively.
The semi-supervised facial landmark detection aims to learn robust detectors from both $\gL$ and $\gU$.

\begin{figure}[t]
\center
\includegraphics[width=\columnwidth]{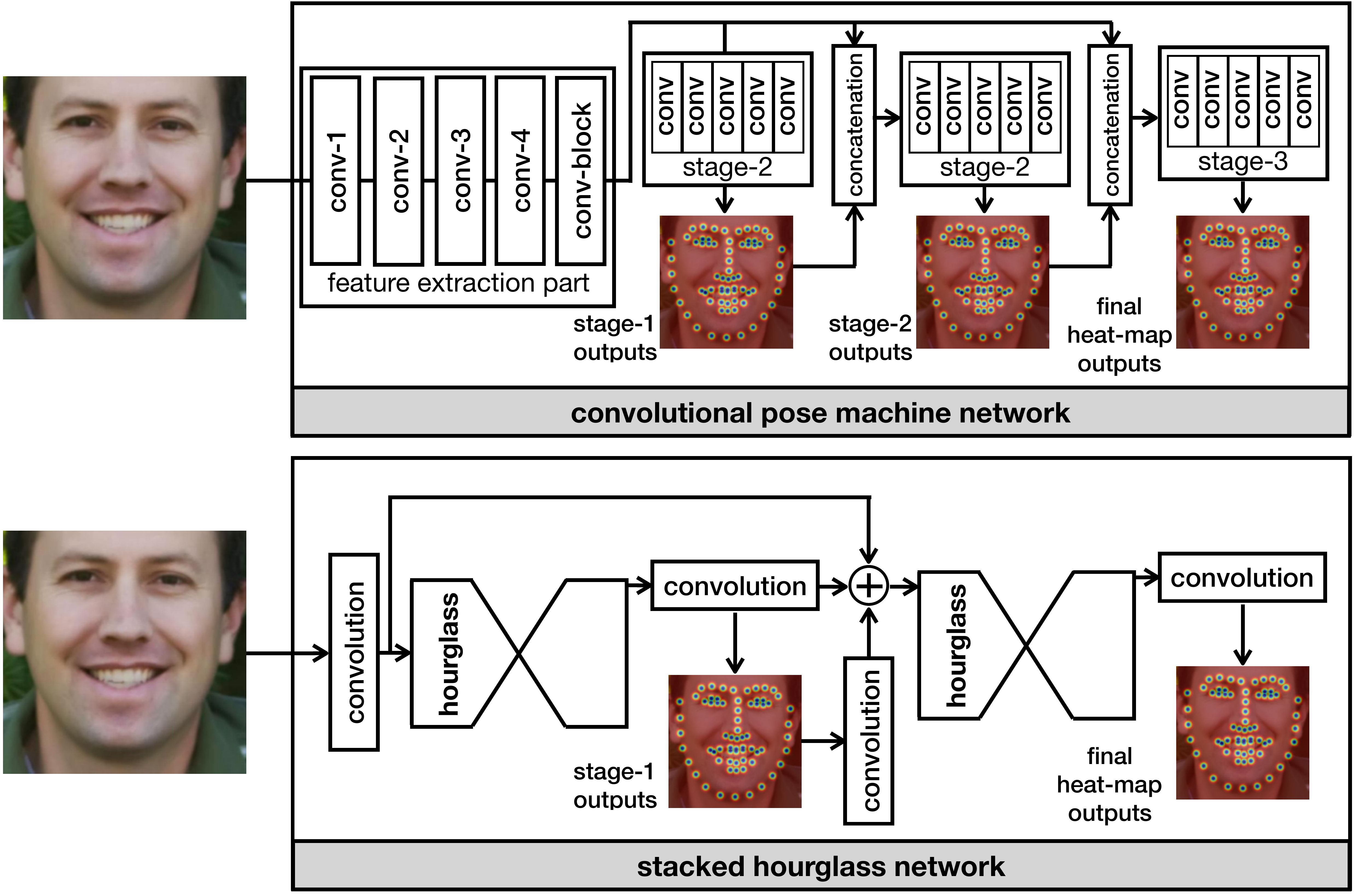}
\caption{
A brief overview of the structure between the two student detection networks in our TS$^{3}$. The first network is convolutional pose machine~\cite{wei2016convolutional} and the second is stacked hourglass~\cite{newell2016stacked}.
}
\label{fig:network}
\end{figure}

\subsection{Teacher and Students Design}\label{sec:teacher-student}

\textbf{The Student Detectors.}
We choose the convolutional pose machine (CPM)~\cite{wei2016convolutional} and stacked hourglass (HG)~\cite{newell2016stacked} models as our student detectors.
These two landmark detection architectures are the cornerstone of many facial landmark detection algorithms~\cite{newell2016stacked,dong2018san,bulat2017far,tang2018quantized}.
Moreover, their architectures are quite different, and can thus complement each other to achieve a better detection performance compared to using two similar neural architectures.
Therefore, we integrate these two detectors in our TS$^{3}$ approach.
In this paragraph, we will give a brief overview of these two facial landmark detectors.
We illustrate the structures of CPM and HG in \Figref{fig:network}. Both CPM and HG are the heatmap regression based methods and utilize the cascaded structure.
Formally, suppose there are $M$ convolutional stages in CPM, the output of CPM is:
{
\begin{align}\label{eq:cpm-hg}
 f_{1}(x_{i} | \vw_{1}) = \{ \mH^{m}_{i} | 1 \leq m \leq M \},
\end{align}
}
\noindent where $f_{1}$ indicates the CPM student detector whose parameters are $\vw_{1}$. $x_{i}$ is the RGB image of the $i$-th data-point and $\mH^{m}_{i} \in \gR^{(K+1) \times h'\times w'}$ indicates the heatmap prediction of the $m$-th stage. $h'$ and $w'$ denote the spatial height and width of the heatmap.
Similarly, we use $f_{2}$ indicates the HG student detector whose parameters are $\vw_{2}$.
The detection loss function of the CPM student is:
{
\begin{align}\label{eq:cpm-hg-loss}
\ell(f_{1}(x_{i} | \vw_{1}), y_{i}) =~&~\sum_{m}^{M} ||\mH^{m}_{i} - \mH^{*}_{i}||_{F}^{2} \nonumber\\
                   =~&~\sum_{m}^{M} ||\mH^{m}_{i} - p(y_{i})||_{F}^{2},
\end{align}
}\noindent where $p$ is a function taking the label $y_{i} \in \gR^{2\times K}$ as inputs to generate the the ideal heatmap $\mH^{*}_{i} \in \gR^{(K+1) \times h'\times w'}$.
Details of $p$ can be found in~\cite{wei2016convolutional,newell2016stacked}. During the evaluation, we take the argmax results over the first $K$ channel of the last heatmap $\mH_{M}$ as the coordinates of landmarks, and the $(K+1)$-th channel corresponding to the background will be omitted.

\begin{figure}[t]
\center
\includegraphics[width=\linewidth]{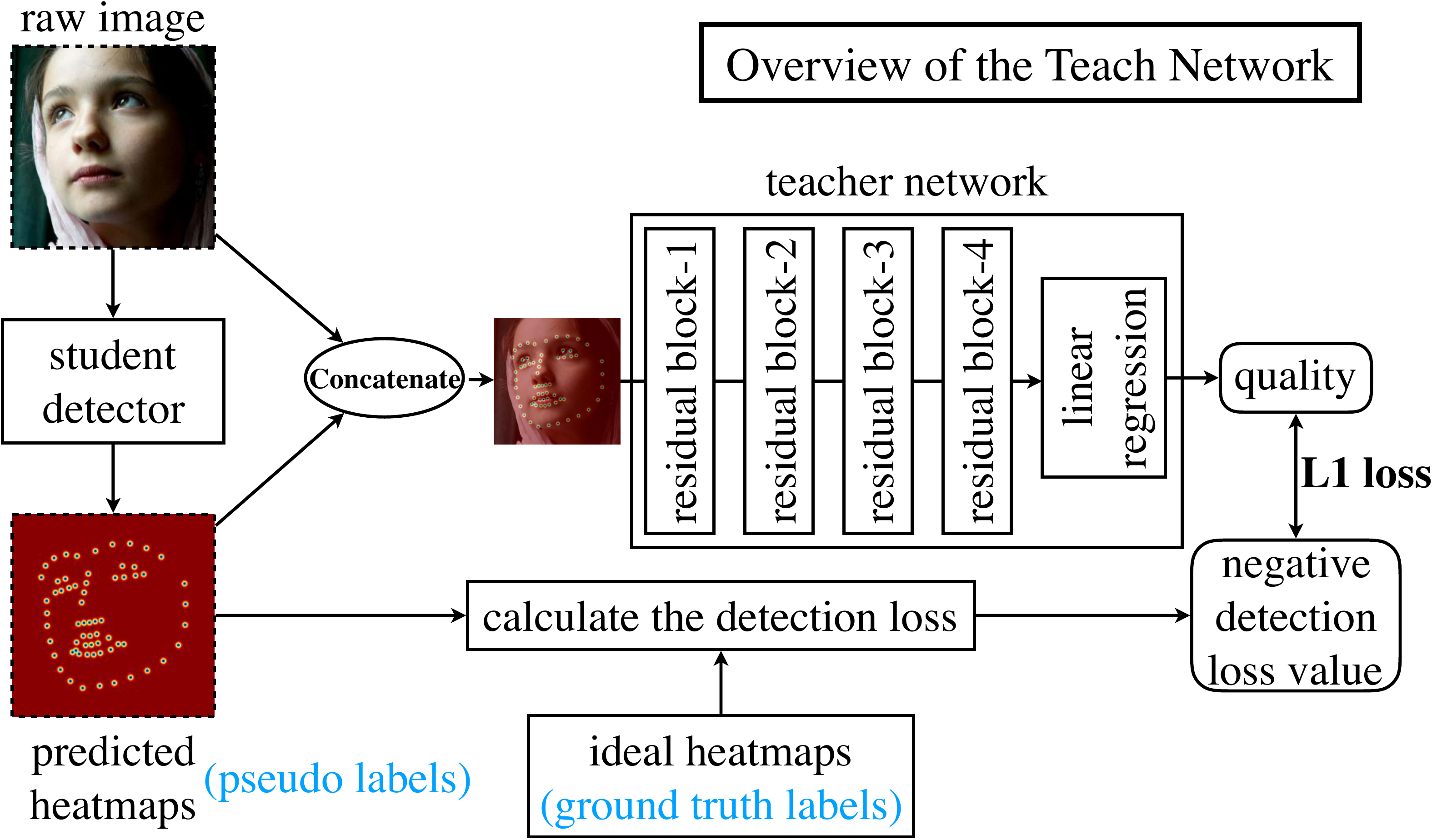}
\caption{
\textbf{The illustration of our teacher network.}
The input of the teacher is the concatenation of the original RGB face image and the heatmap (pseudo label) predicted by the student detector.
The output of teacher is a scalar, representing the quality of the input pseudo labeled face image.
During training, we can calculate a detection loss using the ideal heatmap and the predicted heatmap.
The teacher aims to fit the negative value of this detection loss by an L1 loss.
During evaluation, a higher value of the quality represents a lower detection loss, which means this pseudo labeled image is reliable.
}
\label{fig:value-network}
\end{figure}

\textbf{The Teacher Network.}
Since our student detectors are based on heatmap, the pseudo label is in the form of heatmap and ground truth label is the ideal heatmap.
We build our teacher network using the structure of discriminators adopted in CycleGAN~\cite{zhu2017unpaired}.
As shown in \Figref{fig:value-network}, the input of this teacher network is the concatenation of a face image and its heatmap prediction $\mH^{M}_{i}$\footnote{$\mH^{M}_{i}$ will be resized into the same spatial size as its face image}.
The output of this teacher network is a scalar representing the quality of a pseudo labeled facial image.
Since we train the teacher on the trustworthy labeled data, we could obtain a supervised detection loss by calculating $|| \mH^{M}_{i} - \mH^{*}_{i}||^{2}_{F}$.
We consider the negative value of this detection loss as the ground truth label of the quality, because a high negative value of the detection loss indicates a high similarity between the predicted heatmap and the ideal heatmap. In another word, a higher quality scalar corresponds to a more accurate pseudo label.

Formally, denote the teacher network as $g$, we have:
{
\begin{align}
  g( {x_{i}}^{\frown}\mH^{M}_{i} | \vw_{g}) = {q}_{i} , \label{eq:teacher} \\
  \ell_{t}( g( {x_{i}}^{\frown}\mH^{M}_{i} | \vw_{g}), y_{i} ) = | q + || \mH^{M}_{i} - \mH^{*}_{i} ||_{F}^{2} | , \label{eq:teacher-loss}
\end{align}
}\noindent where the parameters of the teacher is $\vw_{g}$.
``$x^{\frown}\mH$'' first resizes the tensor $\mH$ into the same spatial shape as $x$ and then concatenates the resized tensor with $x$ to get a new tensor. This new tensor is regarded as pseudo labeled image and will be qualified by the teacher later.
The teacher outputs a scalar $q_{i}$ representing the quality of the $i$-th sample associated with its pseudo label $\mH_{i}^{M}$.
We optimize the teacher on the trustworthy labeled data by minimizing \Eqref{eq:teacher-loss}.

\begin{algorithm}[t!]
\caption{The Algorithm Description of Our TS$^{3}$}
\label{alg:SSFLD}
\begin{algorithmic}[1]
\Require Labeled data ${\gL}=\{(x_i,y_i) | 1 \leq i \leq n_{l}\}$ \\
\hspace{4mm} Unlabeled data ${\gU}=\{(x_i^u) | n_{l}+1 \leq i \leq n_{u}+n_{l}\}$ \\
\hspace{4mm} Two student detectors $f_{1}$ with $\vw_{1}$ and $f_{2}$ with $\vw_{2}$ \\
\hspace{4mm} The teacher network $g$ with parameters $\vw_{g}$ \\
\hspace{4mm} The selection ratio $r$ and the maximum step $S$

\State Initialize the $\vw_{1}$ and $\vw_{2}$ by minimizing \Eqref{eq:cpm-hg-loss} on $\gL$

\For{$i=1$; $i \leq S$; $i$++}

  \State Predict $\mH_{i}^{M}$ on both $\gL$ and $\gU$ using \Eqref{eq:ensemble}, and denote $\gU$ with its pseudo labels as $\gU^{1}$ \Comment{update the first student}
  \State Optimize teacher with $\vw_{g}$ by minimizing \Eqref{eq:teacher-loss} on $\gL$ with prediction $\mH_{i}^{M}$ and ground truth label $\mH_{i}^{*}$ 
  \State Compute the quality scalar of each sample in $\gU^{1}$ using the optimized teacher via \Eqref{eq:teacher}
  \State Pickup the top $r\times{i}\times|\gU|$ samples from $\gU^{1}$, named as $\gL_{\textrm{ex}}^{1}$
  \State Retrain $\vw_{1}$ on $\gL^{1} = \gL \cup \gL_{\textrm{ex}}^{1}$ by minimizing \Eqref{eq:cpm-hg-loss}
  
  \State Predict $\mH_{i}^{M}$ on both $\gL$ and $\gU$ using \Eqref{eq:ensemble}, and denote $\gU$ with its pseudo labels as $\gU^{2}$ \Comment{update the second student}
  \State Optimize teacher with $\vw_{g}$ by minimizing \Eqref{eq:teacher-loss} on $\gL$ with $\mH_{i}^{M}$ and $\mH_{i}^{*}$ 
  \State Compute the quality scalar of each sample in $\gU^{2}$ using \Eqref{eq:teacher}
  \State Pickup the top $r\times{i}\times|\gU|$ samples from $\gU^{2}$, named as $\gL_{\textrm{ex}}^{2}$
  \State Retrain $\vw_{2}$ on $\gL^{2} = \gL \cup \gL_{\textrm{ex}}^{2}$ by minimizing \Eqref{eq:cpm-hg-loss}

\EndFor

\Ensure Students with optimized parameters $\vw_{1}$ and $\vw_{2}$

\end{algorithmic}
\end{algorithm}

\subsection{The TS$^{3}$ Algorithm}\label{sec:method-algo}

Our TS$^{3}$ aims to progressively improve the performance of the student detector. The key idea is to learn a teacher network that can teach students which pseudo labeled sample is reliable and can be used for training. In this procedure, we define the pseudo label of a facial image is as follows:
{
\begin{align}\label{eq:ensemble}
    f(x_{i}) = ~&~ \frac{1}{2} (f_{1}(x_{i} | \vw_{1}) + f_{2}(x_{i} | \vw_{2})) \nonumber\\
             = ~&~ \{ \frac{1}{2}(\mH^{(1,m)}_{i} + \mH^{(2,m)}_{i}) | 1 \leq m \leq M \}, \nonumber \\
             = ~&~ \{ \mH^{m}_{i} | 1 \leq m \leq M \},
\end{align}
}\noindent where $\mH^{(1,m)}_{i}$ indicates the heatmap prediction from the first student at the $m$-th stage for the $i$-th sample. $\mH^{m}_{i}$ in \Eqref{eq:ensemble} indicates the ensemble result from both two students detection networks.
It will be used as the prediction during the inference procedure.

We show our overall algorithm in \Algref{alg:SSFLD}.
We first initialize the two detectors $f_{1}$ and $f_{2}$ on the labeled facial images $\gL$.
Then, in the first round, our algorithm applies the following procedures:
(1) generate pseudo labels on $\gL$ via \Eqref{eq:ensemble} and train the teacher network from scratch with these pseudo labels;
(2) generate pseudo labels on $\gU$ and estimate the quality of these pseudo labeled using the learned teacher;
(3) select some high-quality pseudo labeled samples to retrain one student network from scratch.
(4) repeat the first three steps to update another student detection network.
In the next rounds, each student can be improved and generate more accurate pseudo labels. In this way, we will select more pseudo labeled samples when retraining the students.
As the rounds go, students will gradually become better, and the teacher will also be adaptive with the improved students.
Our interaction mechanism helps to obtain more accurate pseudo labels and select more reliable pseudo labeled samples. As a result, our algorithm achieves better performance in the semi-supervised facial landmark detection.\looseness-1

\subsection{Discussion}\label{sec:discussion}

\textbf{Can this algorithm generalize to other tasks?}
Our algorithm relies on the design of the teacher network. It requires the input pseudo label to be a structured prediction.
Therefore, our algorithm is possible to be applied to tasks with structured predictions, such as segmentation and pose estimation, but is not suitable other tasks like classification.

\textbf{Limitation.}
It is challenging for a teacher to judge the quality of a pseudo label for an image, especially when the spatial shape of this image becomes large. Therefore, in this paper, we use an input size of 64$\times$64. If we increase the input size to 256$\times$256, the teacher will fail and need to be modified accordingly.
There are two main reasons: (1) the larger resolution requires a deeper architecture or dilated convolutions for the teacher network and (2) the high-resolution faces bring high-dimensional inputs, and consequently, the teacher needs much more training data.
This drawback limits the extension of our algorithm to high-resolution tasks, such as segmentation. We will explore to solve this problem in the future.

\textbf{Further improvements.}
(1) In our algorithm, during the retraining procedure, a part of unlabeled samples are not involved during retraining.
To utilize these unlabeled facial images, we could use self-supervised techniques such as~\cite{honari2018improving} to improve the detectors.
(2) In this framework, we use only two student detectors, while it is easy to integrate more student detectors. More student detectors are likely to improve the prediction accuracy, but this will introduce more computation costs. 
(3) The specifically designed data augmentation~\cite{radosavovic2018data,wu2017facial} is another direction to improve the accuracy and precision of the pseudo labels.

\textbf{
Will the teacher network over-fit to the labeled data?
}
In \Algref{alg:SSFLD}, since labeled data set $\gL$ is used to optimize both teacher and students, the teacher's judgment could suffer from the over-fitting problem.
Most of the students' predictions on the labeled data can be similar to the ground truth labels. In other words, most pseudo labeled samples on $\gL$ are ``correctly'' labeled samples.
If the teacher is optimized on $\gL$ with those pseudo labels, it might only learn what a good pseudo labeled sample is, but overlook what a bad one is.
It would be more reasonable to let students predict on the unseen validation set, and then train the teacher on this validation set.
However, having an additional validation set during training is different from the typical setting of previous semi-supervised facial landmark detection. We would explore this problem in our future work.


\section{Empirical Studies}\label{sec:experiments}

We perform experiments on three benchmark datasets to investigate the behavior of the proposed method.
The datasets and experiment settings are introduced in \Secref{sec:datasets} and \Secref{sec:setting}.
We first compare the proposed semi-supervised facial landmark algorithm with other state-of-the-art algorithms in Sec.~\ref{sec:compare}.
We then perform ablation studies in Sec.~\ref{sec:ablation} and visualize our results at last.

\begin{table}[!t]
\setlength{\tabcolsep}{2pt}
\centering
\begin{tabular}{|c|c|c|c|c|} \hline\hline
 Ratio &      Method                                  & Common   & Challenging &  Full   \\\hline
 100\% &      MDM \cite{trigeorgis2016mnemonic}       & 4.83     & 10.14       &  5.88      \\
 100\% &      Two-Stage \cite{lv2017deep}             & 4.36     & 7.42        &  4.96      \\
 100\% &      RDR \cite{xiao2017recurrent}            & 5.03     & 8.95        &  5.80      \\ 
 100\% & Pose-Invariant~\cite{jourabloo2017pose}      & 5.43     & 9.88        &  6.30      \\
 100\% & HF-ResNet~\cite{ranjan2019hyperface}         &  -       & 8.18        & - \\
 100\% &      SAN \cite{dong2018san}                  & 3.34     & 6.60        &  3.98      \\
 100\%$^{\dag}$ &  SBR \cite{dong2018sbr}             & 3.28     & 7.58        &  4.10      \\
 100\% &      PCD-CNN \cite{kumar2018disentangling}   & 3.67     & 7.62        &  4.44      \\\hline\hline
 10\%  & RCN$^{+}$~\cite{honari2018improving}         &   -      & 10.35       &  6.32      \\
 10\%  &      {\NAME}                                 & {4.67}   & {9.26}      &  {5.64}    \\\hline
 20\%  & RCN$^{+}$~\cite{honari2018improving}         &   -      & 9.56        &  5.88      \\
 20\%  &      {\NAME}                                 & {4.31}   & {7.97}      &  {5.03}    \\\hline
 100\%       &      {\NAME}                           & 3.17     & 6.41        &  3.78    \\\hline
 100\%$^{\ddagger}$&      {\NAME}                      & 2.91     & 5.91        &  3.49    \\\hline
 \hline
\end{tabular}
\vspace{2mm}
\caption{
Comparisons of the NME results on the 300-W dataset. ``Ratio'' indicates the annotation ratio of the whole training set.
A ``Ratio'' value of 10\% means that only 10\% of the training face images have the landmark coordinate labels.
$^{\dag}$ indicates that SBR \cite{dong2018sbr} used additional unlabeled video data during training.
When we use partially labeled training images, our {\NAME} outperforms other semi-supervised algorithm~\cite{honari2018improving}.
$^{\ddagger}$ indicates we use 100\% labeled 300-W training data and unlabeled AFLW training data for our {\NAME}.
}
\label{table:300W-ALL}
\end{table}

\subsection{Datasets}\label{sec:datasets}

\textbf{The 300-W dataset}~\cite{sagonas2013300} annotates 68 landmarks from five facial landmark datasets, i.e., LFPW, AFW, HELEN, XM2VTS, and IBUG.
Following the common settings~\cite{dong2018sbr,dong2018san,lv2017deep}, we regard all the training samples from LFPW, HELEN and the full set of AFW as the training set, in which there is 3148 training images.
The common test subset consists of 554 test images from LFPW and HELEN.
The challenging test subset consists of 135 images from IBUG to construct .
The full test set the union of the common and challenging subsets, 689 images in total.

\begin{table*}[t]
\setlength{\tabcolsep}{6pt}
\centering
\begin{tabular}{|c|c c c c c c c|} \hline\hline
Methods    & SDM~\cite{xiong2013supervised} & LBF~\cite{ren2016face} & CCL~\cite{zhu2016unconstrained} & Two-Stage~\cite{lv2017deep} & SBR~\cite{dong2018san}$\dag$ & SAN~\cite{dong2018san}   & DSRN~\cite{miao2018direct} \\ \hline
AFLW-Full  & 4.05                           & 4.25                   & 2.72                            & 2.17                        & 2.14                   & 1.91   & 1.86 \\ \hline
AFLW-Front & 2.94                           & 2.74                   & 2.17                            & -                           & 2.07                   & 1.85  & - \\\hline\hline
Methods    & RCN$^{+}$~\cite{honari2018improving} (5\%) & \multicolumn{2}{c}{{\NAME} (5\%)} & \multicolumn{2}{c}{{\NAME}(10\%)} & \multicolumn{2}{c|}{{\NAME}(20\%)} \\\hline
AFLW-Full  & 2.17                                       & \multicolumn{2}{c}{2.19}       & \multicolumn{2}{c}{2.14} & \multicolumn{2}{c|}{1.99}\\\hline
AFLW-Front & -                                          & \multicolumn{2}{c}{2.03}       & \multicolumn{2}{c}{1.94} & \multicolumn{2}{c|}{1.86} \\\hline\hline
\end{tabular}
\vspace{2mm}
\caption{
Comparisons of NME normalized by face size on the AFLW dataset.
$\dag$ indicates that SBR~\cite{dong2018sbr} used additional unlabeled video data during training.
The ratio number in the brackets represents the portion of the labels that we use. Compared to the semi-supervised algorithm~\cite{honari2018improving}, our {\NAME} obtains a similar NME result (2.19 vs. 2.17).
Compared to supervised algorithms which use 100\% labels, our {\NAME} obtains competitive NME when using only 20\% labels.
}
\label{table:aflw}
\end{table*}

\textbf{The AFLW dataset}~\cite{koestinger2011annotated} contains 21997 real-world images with 25993 faces in total.
They provide at most 21 landmark coordinates for each face, but they exclude invisible landmarks.
Faces in AFLW usually have a different head pose, expression, occlusion or illumination, and therefore it causes difficulties to train a robust detector.
Following the same setting as in \cite{lv2017deep,zhu2016unconstrained}, we do not use the landmarks of two ears.
There are two types of AFLW splits, i.e., AFLW-Full and AFLW-Frontal following~\cite{zhu2016unconstrained,dong2018san}.
AFLW-Full contains 20000 training samples and 4386 test samples.
AFLW-Front uses the same training samples as in AFLW-Full, but only use the 1165 samples with the frontal face as the test set.

\textbf{The 300-VW dataset}~\cite{shen2015first} is a video-based facial landmark benchmark. It contains 50 training videos with 95192 frames. Following~\cite{khan2017synergy,dong2018sbr}, we report the results for the 49 inner points on the category C subset of the 300-VW test set, which has 26338 frames.

\subsection{Experimental Settings}\label{sec:setting}

\textbf{Training student detection networks.}
The first student detector is CPM~\cite{wei2016convolutional}.
We follow the same model configuration as the base detector used in~\cite{wei2016convolutional,dong2018san}, and the number of cascaded stages is set as three.
Its number of parameters is 16.70 MB and its FLOPs is 1720.98 M.
To train CPM, we apply the SGD optimizer with the momentum of 0.9 and the weight decay of 0.0005.
For each stage, we train the CPM for 50 epochs in total.
We start the learning rate of 0.00005, and reduce it by 0.5 at 20-th, 25-th, 30-th, and 40-th epoch.

The second student detector is HG~\cite{newell2016stacked}.
We follow the same model configuration as~\cite{bulat2017far} but use the number of cascaded stages of four to build our HG model, where the number of parameters is 24.97 MB and FLOPs is 1600.85 M.
To train HG, we apply the RMSprop optimizer with the alpha of 0.99.
For each stage, we train the HG for 110 epochs in total.
We start the learning rate of 0.00025, and reduce it by 0.5 at 50-th, 70-th, 90-th, and 100-th.

For both of these two detectors, we use the batch size of eight on two GPUs.
To generate the heatmap ground truth labels, we apply the Gaussian distribution with the sigma of 3.
Each face image is first resized into the size of 64$\times$64, and then randomly resized between the scale of 0.9 and 1.1.
After the random resize operation, the face image will be randomly rotated with the maximum degree of 30, and then randomly cropped with the size of 64$\times$64\footnote{Different input image resolution can cause different detection performance. We choose 64$\times$64 to ease the training of our teacher network.}.
We set selection ratio $r$ as $0.1$ and the maximum step $S$ as $6$ based on cross-validation.

\textbf{Training the teacher network\footnote{Model codes are publicly available on GitHub: {https://github.com/D-X-Y/landmark-detection}}.}
We build our teacher network using the structure of discriminators adopted in CycleGAN~\cite{zhu2017unpaired}.
Given a 64$\times$64 face image, we first resize the predicted heatmap into the same spatial size of 64$\times$64.
We use the Adam to train this teacher network.
The initial learning rate is 0.01, and the batch size is 128.
Random flip, random rotation, random scale and crop are applied as data argumentation.

\textbf{Evaluation.}
Normalized Mean Error (NME) is usually applied to evaluate the performance for facial landmark predictions~\cite{lv2017deep,ren2016face,zhu2016unconstrained,dong2018san}.
For the 300-W dataset, we use the inter-ocular distance to normalize mean error following the same setting as in~\cite{sagonas2013300,lv2017deep,dong2018sbr,dong2018san}.
For the AFLW dataset, we use the face size to normalize mean error~\cite{lv2017deep}.
Area Under the Curve (AUC) @ 0.08 error is also employed for evaluation~\cite{bulat2017far,trigeorgis2016mnemonic}.
When training on the partially labeled data, the sets of $\gL$ and $\gU$ are randomly sampled.
During evaluation, we use \Eqref{eq:ensemble} to obtain the final heatmap and follow~\cite{wei2016convolutional,newell2016stacked} to generate the coordinate of each landmark.
We repeat each experiment three times and report the mean result.
The codes will be public available upon the acceptance.

\begin{table}[t]
\setlength{\tabcolsep}{8pt}
\centering
\begin{tabular}{|c|c|c|c|c|} \hline\hline
 Method     & DGCM~\cite{khan2017synergy}& SBR~\cite{dong2018sbr} & {\NAME}  \\\hline
AUC@0.08    &  59.38                     & 59.39 &  59.65 \\
\hline\hline
\end{tabular}
\vspace{2mm}
\caption{
AUC @ 0.08 error on 300-VW category C.
Note that all compared algorithms~\cite{khan2017synergy,dong2018sbr} use all labels on the 300-VW training data and 300-W training data, whereas our {\NAME} only uses the unlabeled 300-VW training data and labeled 300-W training data.
}
\label{table:300VW-C}
\end{table}

\subsection{Comparison with state-of-the-art}\label{sec:compare}

\textbf{Comparisons on 300-W.}
We compare our algorithm with several state-of-the-art algorithms~\cite{xiong2013supervised,xiao2017recurrent,lv2017deep,xiao2017recurrent,jourabloo2017pose,honari2018improving}, as shown in \Tabref{table:300W-ALL}.
In this table, \cite{dong2018san,kumar2018disentangling,dong2018sbr} are very recent methods, which represent the state-of-the-art supervised facial landmark algorithms.
By using 100\% facial landmark labels on 300-W training set and unlabeled AFLW, our algorithm achieves competitive 3.49 NME on the 300-W common test set, which is competitive to other state-of-the-art algorithms.
In addition, even though our approach utilizes two detectors, the number of parameters is much lower than SAN~\cite{dong2018san}.
The robust detection performance of ours can be mainly caused by two reasons.
First, the proposed teacher network can effectively sample the qualified pseudo labeled data, which enables the model to exploit more useful information.
Second, our framework leverages two advanced CNN architectures, which can complement each other.

We also compare our {\NAME} with a recent work on semi-supervised facial landmark detection~\cite{honari2018improving} in \Tabref{table:300W-ALL}.
When using 10\% of labels, our {\NAME} obtains a lower NME result on the challenging test set than RCN$^{+}$~\cite{honari2018improving}~(5.64 NME \emph{vs.} 6.32 NME).
When using 20\% of labels, our {\NAME} is also superior to it (5.03 NME \emph{vs.} 5.88 NME).
Note that \cite{honari2018improving} utilizes a transformation invariant auxiliary loss function.
This auxiliary loss can also be easily integrated into our framework.
Therefore, \cite{honari2018improving} is orthogonal to our work, combining two methods can potentially achieve a better performance.

\textbf{Comparisons on AFLW.}
We also show the NME comparison on the AFLW dataset in \Tabref{table:aflw}.
Compared to semi-supervised facial landmark detection algorithm~\cite{honari2018improving}, we achieve a similar performance.
RCN$^{+}$~\cite{honari2018improving} can learn transformation invariant information from a large amount of unlabeled images, while ours does not consider this information as it is not our focus.
On the AFLW-Full test set, using 20\% annotation, our framework achieves 1.99 NME, which is competitive to other supervised algorithms.
On the  AFLW-Front test set, using only 10\% annotation, our framework achieves competitive NME results to~\cite{dong2018san}.
The above results demonstrate our framework can train a robust detector with much less annotation effort.

\begin{figure}[t]
\center
\includegraphics[width=\columnwidth]{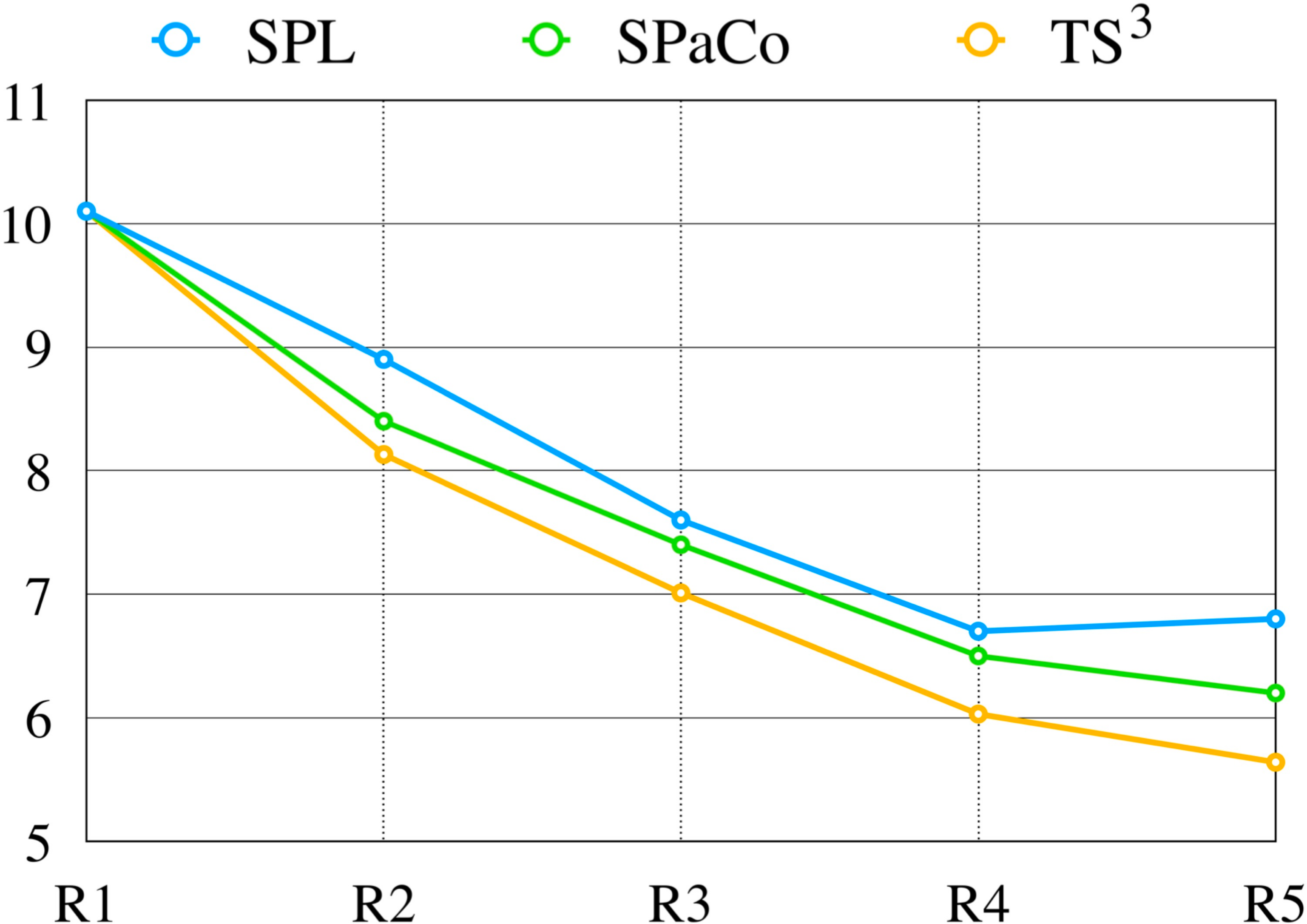}
\caption{
We compare three different algorithms, which can train two detectors in a progressive manner: SPL~\cite{kumar2010self,jiang2015self}, SPaCo~\cite{ma2017self}, and our {\NAME}.
All these algorithms iteratively improve detectors one round by another round. The x-axis shows the results of the first five rounds. The y-axis indicates the NME results on the 300-W full test set.
}
\label{fig:compare}
\end{figure}

\textbf{Comparisons on 300-VW.}
We experiment our algorithm to leverage a large amount of unlabeled facial video frames on 300-VW.
We use the labeled 300-W training set and the unlabeled 300-VW training set to train our {\NAME}.
We evaluate the learned detectors on the 300-VW C test subset w.r.t. AUC~@~0.08.
Some video-based facial landmark detection algorithms~\cite{khan2017synergy,dong2018sbr} utilize the labeled 300-VW training data to improve the base detectors.
Compared with them, without using any label on 300-VW, our {\NAME} obtains a higher AUC result than them, i.e., 59.65 vs. 59.39, as shown in \Tabref{table:300VW-C}.

\subsection{Ablation Study}\label{sec:ablation}

The key contribution of our {\NAME} lies on two components:
(1) the teacher supervising the training data selection of students.
(2) the complementary effect of two students.
In this subsection, we validate the contribution of these two components to the final detection performance.

\textbf{The effect of the teacher.}
Compared to other progressive pseudo label generation strategies~\cite{kumar2010self,jiang2015self,ma2017self}, our designed teacher can sample pseudo labeled with higher quality.
In \Figref{fig:compare}, we show the detection results after the first five training rounds (only 10\% labels are used).
We use SPL~\cite{kumar2010self,jiang2015self} to separately train CPM and HG, and then ensemble them together as \Eqref{eq:ensemble}.
We use SPaCo~\cite{ma2017self} to jointly optimize CPM and HG in a co-training strategy.
To make a fair comparison, at each round, we control the number of pseudo labels is the same across these three algorithms.
From \Figref{fig:compare}, several conclusions can be made:
(1) {\NAME} obtains the lowest NME, because the quality of selected pseudo labels is better than others.
(2) SPL falls into a local trap at round$_{4}$ and results in a higher error at round$_{5}$, whereas SPaCo and our {\NAME} not. This could be caused by that the interaction between two students can help regularize each other.
(3) Our {\NAME} converges faster than SPaCo and achieves better results. The pseudo labeled data selection in SPaco is a heuristic unsupervised criterion, whereas our criterion is a supervised teacher. Since no extra supervision is given in SPaCo, their criterion might induce inaccurate pseudo labeled samples.
Besides, as discussed in \Secref{sec:discussion}, our {\NAME} can utilize validation set to further improve the performance by avoid over-fitting, but the compared methods may not effectively utilize validation set.

\begin{table}[t]
\setlength{\tabcolsep}{2.5mm} 
\centering
\begin{tabular}{|c|c|c|c|c|} \hline\hline
   Ratio               & Method              & Common & Challenging & Full \\\hline
 \multirow{3}{*}{10\%} & CPM                 & 6.86   & 14.69       & 8.28     \\
                       & HG                  & 5.16   & 11.28       & 6.25     \\
                       & {\NAME}               & 4.67   & 9.26        & 5.64     \\\hline

 \multirow{3}{*}{20\%} & CPM                 & 5.36   & 11.31       & 6.68     \\
                       & HG                  & 5.84   & 10.15       & 6.68     \\
                       & {\NAME}               & 4.31   & 7.97        & 5.03     \\\hline\hline
\end{tabular}
\vspace{2mm}
\caption{
Comparisons of the NME results on the 300-W test sets for different configuration and models.
CPM and HG indicate using only one CPM student or only one HG student in our framework.
When using a single detector, we use the heatmap of the last stage in \Eqref{eq:cpm-hg} as prediction.
When using two students ({\NAME}), we use $\mH_{i}^{M}$ in \Eqref{eq:ensemble} as prediction.
``Ratio'' indicates the proportion of labeled data in our semi-supervised setting.
}
\vspace{-2mm}
\label{table:300W-ALL-Ablation}
\end{table}

\textbf{The effect of the interaction between students.}
From Table~\ref{table:300W-ALL-Ablation}, we show the ablative studies on the complementary effect of multiple students.
In these experiments, we use the same teacher structure, while ``CPM'' and ``HG'' are trained without the interaction between students.
Using 10\% labels, CPM achieves 8.28 NME, and HG achieves 6.25 NME on 300-W.
Leveraging from their mutual benefits, our {\NAME} can boost the performance to 5.64, which is higher than CPM by about 30\% and than HG by 9\%.
Under different portion of annotations, we can conclude similar observations. This ablation study demonstrates the contribution of student interaction to the final performance. 
Note that, our algorithm can be readily applied to multiple students without introducing additional hyper-parameters.
In contrast, the number of hyper-parameters in other co-training strategies~\cite{ma2017self,dong2018few} is quadratic to the number of detectors.

\subsection{Qualitative Analysis}

\begin{figure}[t]
\center
\includegraphics[width=\linewidth]{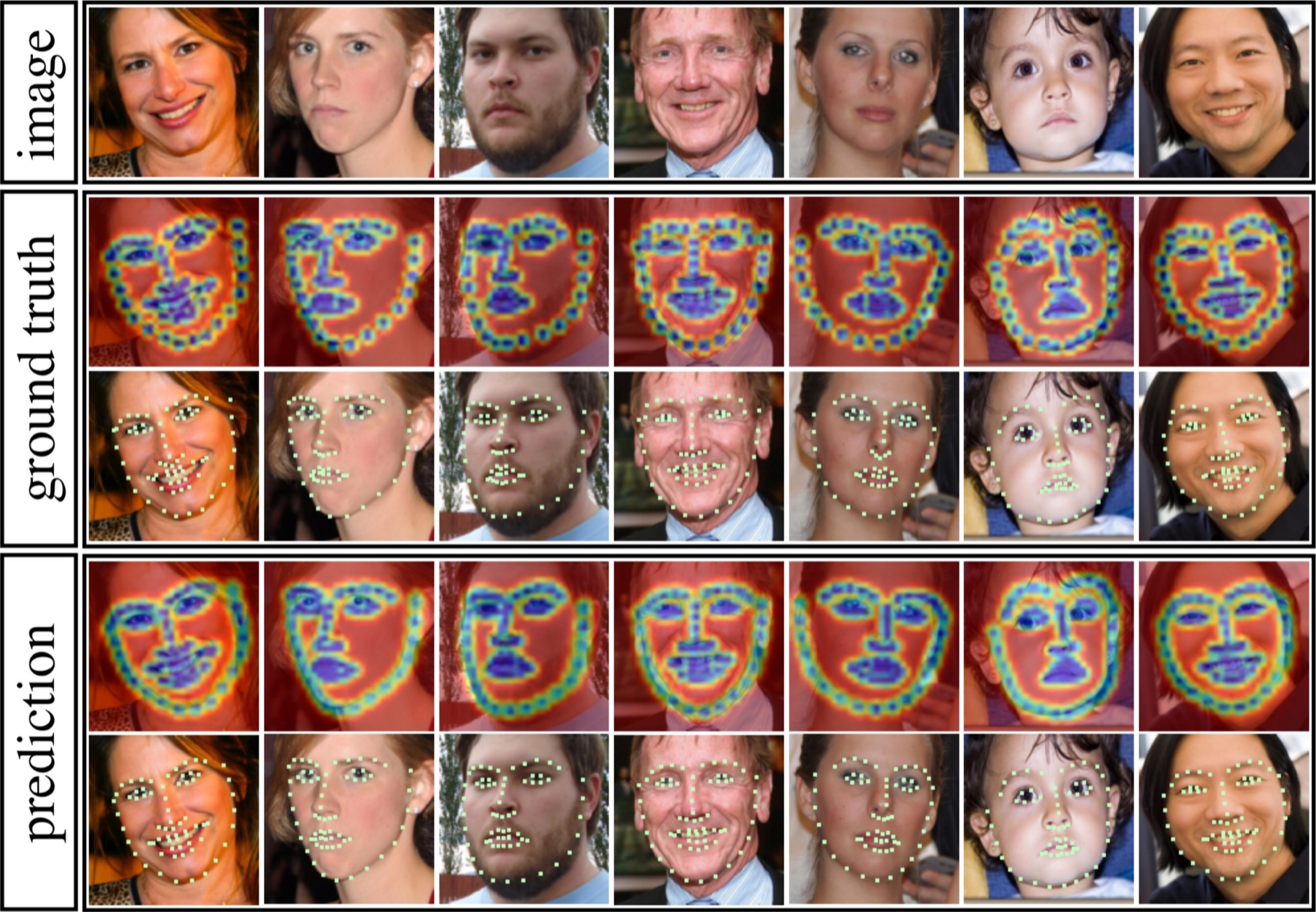}
\caption{
\textbf{Qualitative results on images in the 300-W test set.}
We train our {\NAME} with 314 labeled facial images and 2834 unlabeled facial images in the 300-W training set.
}
\vspace{-2mm}
\label{fig:results}
\end{figure}

On the 300-W training set, we train our {\NAME} using only 10\% labeled facial images, and we show some qualitative results of the 300-W test set in \Figref{fig:results}.
The first row shows seven raw input facial images.
The second row shows the ground truth background heatmaps, and the third row shows the faces with ground truth landmarks of these images.
We visualize the predicted background heatmap in the fourth row and the predicted coordinates in the fifth row.
As we can see, the predicted landmarks of our {\NAME} are very close to the ground truth.
These predictions are already robust enough, and human may not be able to distinguish the difference between our predictions (the third line) and the ground truth (the fifth line).

\section{Conclusion}

In this paper, we propose an interaction mechanism between a teacher and multiple students for semi-supervised facial landmark detection. The students learn to generate pseudo labels for the unlabeled data, while the teacher learns to judge the quality of these pseudo labeled data. After that, the teacher can filter out unqualified samples; and the students get feedback from the teacher and improve itself by the qualified samples. The teacher is adaptive along with the improved students. Besides, multiple students can not only regularize each other but also be ensembled to predict more accurate pseudo labels. We empirically demonstrate that the proposed interaction mechanism achieves state-of-the-art performance on three facial landmark benchmarks.

{\small
\bibliographystyle{ieee_fullname}
\bibliography{egbib}
}

\end{document}